\documentclass{article}

\usepackage[final, nonatbib]{neurips_2023}

\usepackage[utf8]{inputenc} 
\usepackage[T1]{fontenc}    
\usepackage{hyperref}       
\usepackage{url}            
\usepackage{booktabs}       
\usepackage{amsfonts}       
\usepackage{nicefrac}       
\usepackage{microtype}      
\usepackage{xcolor}         
\usepackage{graphicx}
\usepackage{caption}
\usepackage{subcaption}
\usepackage{tabularx}
\usepackage{xcolor}
\usepackage{colortbl}
\definecolor{lightblue}{rgb}{0.88,0.95,1}
\usepackage{adjustbox}
\usepackage{multirow}

\title{Stable Diffusion For Aerial Object Detection}

\author{%
Yanan Jian \\
Microsoft Corp \\
\texttt{yananjian@microsoft.com}
\And 
Fuxun Yu \\
Microsoft Corp\\
\texttt{fuxunyu@microsoft.com} \\
\AND
Simranjit Singh \\
Microsoft Corp\\
\texttt{simsingh@microsoft.com} \\
\And
Dimitrios Stamoulis \\
Microsoft Corp\\
\texttt{distamo@microsoft.com} \\
}

\begin{document}

\maketitle

\begin{abstract}
Aerial object detection is a challenging task, in which one major obstacle lies in the limitations of large-scale data collection and the long-tail distribution of certain classes. Synthetic data offers a promising solution, especially with recent advances in diffusion-based methods like stable diffusion (SD). However, the direct application of diffusion methods to aerial domains poses unique challenges: stable diffusion's optimization for rich ground-level semantics doesn't align with the sparse nature of aerial objects, and the extraction of post-synthesis object coordinates remains problematic. To address these challenges, we introduce a synthetic data augmentation framework tailored for aerial images. It encompasses sparse-to-dense region of interest (ROI) extraction to bridge the semantic gap, fine-tuning the diffusion model with low-rank adaptation (LORA) to circumvent exhaustive retraining, and finally, a Copy-Paste method to compose synthesized objects with backgrounds, providing a nuanced approach to aerial object detection through synthetic data.

\end{abstract}

\section{Introduction}

Aerial imagery \cite{9560031}, derived from various sources such as drones, satellites, and high-altitude platforms, plays a pivotal role in numerous real-world applications. One of the key tasks is object detection which aims to identify and locate objects of interest within a broader scene. However, the path to robust object detection in aerial images is challenging. Unlike ground-level images that are widely available, aerial data are restricted in dataset volume and generalization across different object classes~\cite{9560031}. An even more pressing concern is the long-tail distribution observed in these datasets, where certain classes of objects are vastly underrepresented. This sparsity of examples and the rare classes poses a significant hurdle for the training of generalized object detectors, as models often struggle to recognize and accurately detect these infrequent objects in real-world scenarios. 

To mitigate the scarcity and long-tail distribution of aerial data, synthetic data has emerged as a promising solution~\cite{ghiasi2021simple, goodfellow2014generative, Rombach_2022_CVPR}. Over the years, a plethora of synthetic techniques has been proposed, from the Copy-Paste method~\cite{ghiasi2021simple} where objects from existing images are superimposed onto different backgrounds to simulate new scenes, to advanced DNN-based image synthesis such as Generative Adversarial Networks (GANs)~\cite{goodfellow2014generative}. Recently, synthetic image generation has witnessed major advances from Diffusion-based techniques~\cite{Rombach_2022_CVPR}, which progressively improve an initial noisy image through a series of diffusion steps, and are able to generate photo-realistic natural or artistic images. 

Although undeniably promising, applying diffusion-based synthetic methods to aerial object detection isn't straightforward. Specifically, we observe two major challenges: 
\begin{itemize}
    \item Sparse Region-of-Interests (ROIs): Diffusion methods have primarily been optimized for ground-level images abundant in semantic richness, while aerial detection often features object regions that are sparsely situated. For instance, while aerial urban scenes might be bustling with roads and buildings, aerial detectors might predominantly capture isolated cars or sparse ships traversing vast river expanses. This disparity in object density and semantics necessitates a tailored approach for diffusing aerial scenes.
    \item Coordinate Extraction: While synthesizing whole images might be feasible with diffusion methods, a subsequent challenge arises for detection scenarios: how to extract the precise coordinates of the synthesized objects? Without this critical information, the synthetic images, no matter how realistic, may fall short in their utility for detection tasks.
\end{itemize}

To address these concerns, we introduce a novel framework tailored to the specificities of aerial images with the following steps: 
\begin{itemize}
    \item Sparse-to-Dense ROI extraction: We extract semantically meaningful object patches from sparse aerial scenes, aiming to mirror the rich semantics typical of ground-level images. Paired with text prompts, diffusion models thus could be effectively trained to generate semantic meaningful object patches.
    \item Fine-tuning using LORA: Training a foundation diffusion model from scratch is computing prohibitive. We thus leverage LORA (Low-Rank Adaptation)~\cite{hu2021lora} to fine-tune the off-the-shelf stable diffusion model that circumvents the need for heavy full-model tuning.
    \item Copy-Paste Composition: Finally, synthesized object patches are seamlessly integrated into real-world aerial background images using the Copy-Paste method~\cite{ghiasi2021simple}, where coordinates can be easily obtained and facilitate the downstream object detection task.

\end{itemize}

Applied on the representative aerial object detection dataset DOTAv2.0~\cite{9560031}, our framework demonstrates consistent improvements on both overall and long-tailed classes, e.g., \textbf{+1.2\%} to \textbf{+2.7\%} mean mAP improvement, and at most \textbf{+30.3\%} mAP improvement on extremely long-tail classes.

\section{Methodology}
\label{headings}

\paragraph{Overview of Model/Pipeline}

\begin{figure}[!b]
\vspace{-5mm}
    \centering
    \includegraphics[width=\textwidth]{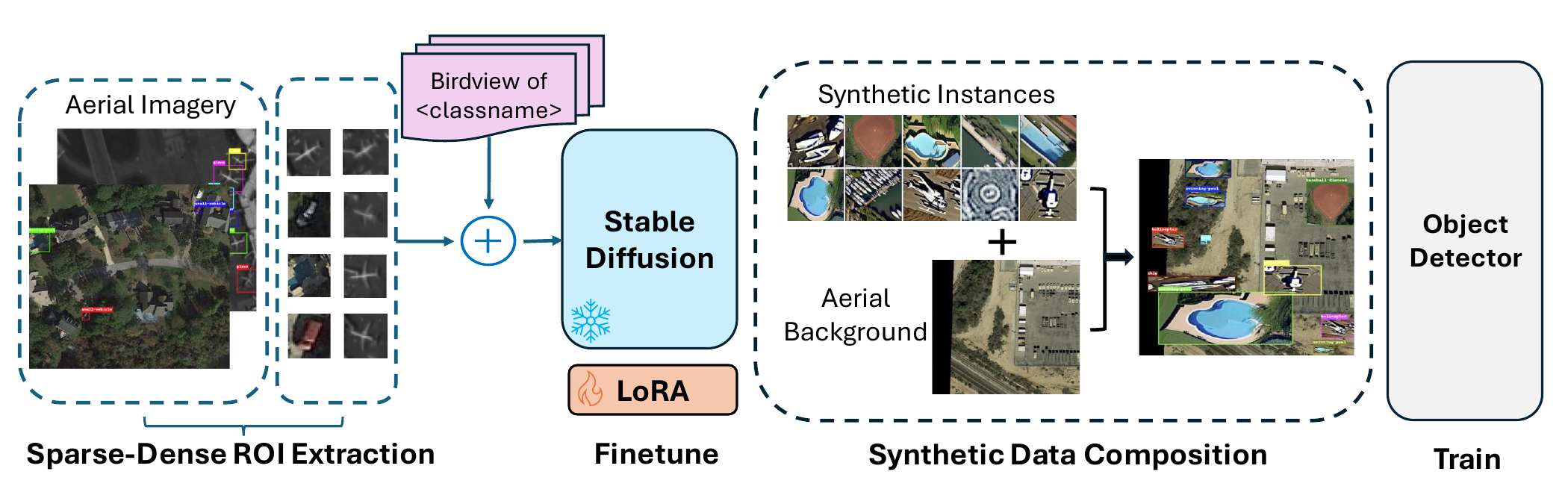}
    \caption{Overview of our proposed approach.}
    \label{fig:mesh1}
\end{figure}

Figure~\ref{fig:mesh1} shows the overall pipeline of our proposed method, which composes three steps to generate the aerial synthetic data. 
(1) \textit{Sparse-to-Dense ROI Extraction}: Given the targeted aerial dataset, we first crop the ground-truth boxes from the object detection training dataset, and add a text prompt label for each of the crops following the format of "birdview of <class name>".
(2) \textit{Stable Diffusion Finetuning with LORA}: We then sample the same number of crops from each class following step 1, which alleviates the potential effect of long-tail impact for stable diffusion (SD) training. The paired object patch and text prompts are then used to finetune SD with low-rank adaptation (LORA) that enables efficient stable diffusion finetuning. 
(3) \textit{Copy-Paste Composition}: Once the SD model is finetuned, we can generate synthetic objects by inferencing SD with different seeds and the text prompt of "birdview of <class name>" for every class. The generated crops are then pasted onto the real aerial backgrounds with the object coordinates extracted automatically. Unlimited synthetic data could be thus generated for training augmentation.

\subsection{Sparse-to-Dense ROI Extraction}
\begin{figure}[!t]
    \centering
    \includegraphics[width=\textwidth]{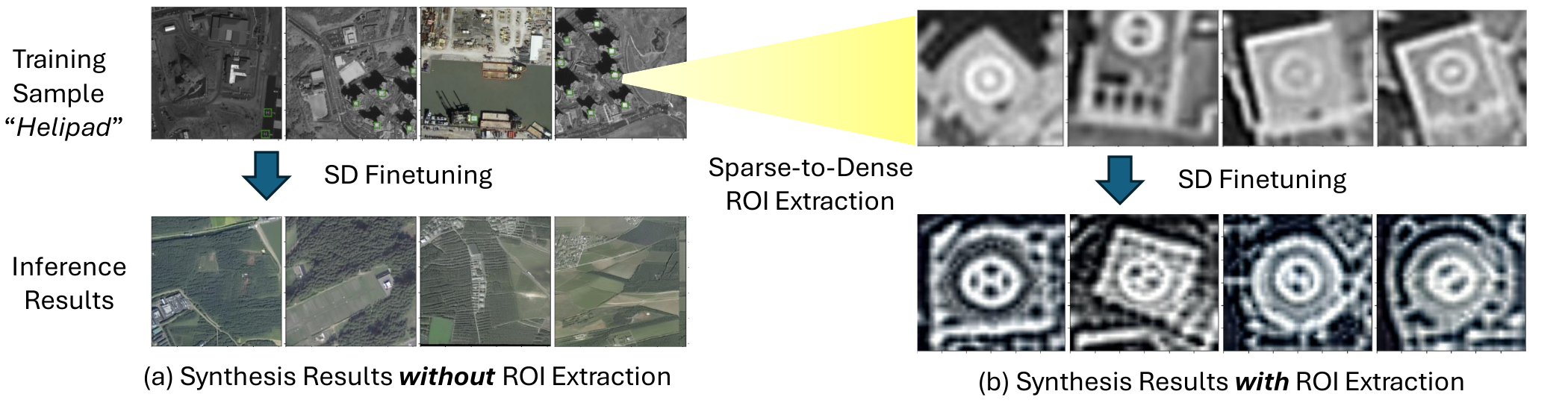}
    \caption{Sparse-to-Dense ROI Extraction is the key for stable diffusion (SD) to learn \textit{meaningful} semantics of targeted objects, without which SD tends to generate empty background images only.}
    \label{fig:sparse_to_dense}
\vspace{-5mm}
\end{figure}

Aerial imagery has a unique attribute of sparse semantic distribution that is different from ground-level natural images - especially regarding the tiny object sizes within the wide-range scenes, as shown in Figure~\ref{fig:sparse_to_dense}. Meanwhile, one single aerial image can contain multiple object classes. Although we can concat multiple class names into the prompts, the sparse semantics make the diffusion training process extremely unstable since it's hard for image encoders to correctly map the region-level semantics with the corresponding texts. As a result, finetuned SD models will usually yield realistic but empty background images without targeted object classes.

To address such a challenge, we propose a sparse-to-dense ROI extraction as a pre-processing step for SD finetuning data preparation. Specifically, we crop out all ground-truth object patches and pair each object with a text prompt as "birdview of <classname>", which ensures the object patch semantics are densely and correctly mapped to the text prompts. We also enlarge the object margin by 10 pixels to make sure we capture enough context. We then use the cropped-out patches and the corresponding text prompt as SD's finetuning data. Figure~\ref{fig:sparse_to_dense} compares the SD synthesis results with/without ROI extraction. Without this step, SD models can hardly learn any meaningful concepts due to the sparse semantics.

\subsection{Stable Diffusion Finetuning with LORA}
\begin{figure}[!b]
\vspace{-3mm}
    \centering
    \includegraphics[width=\textwidth]{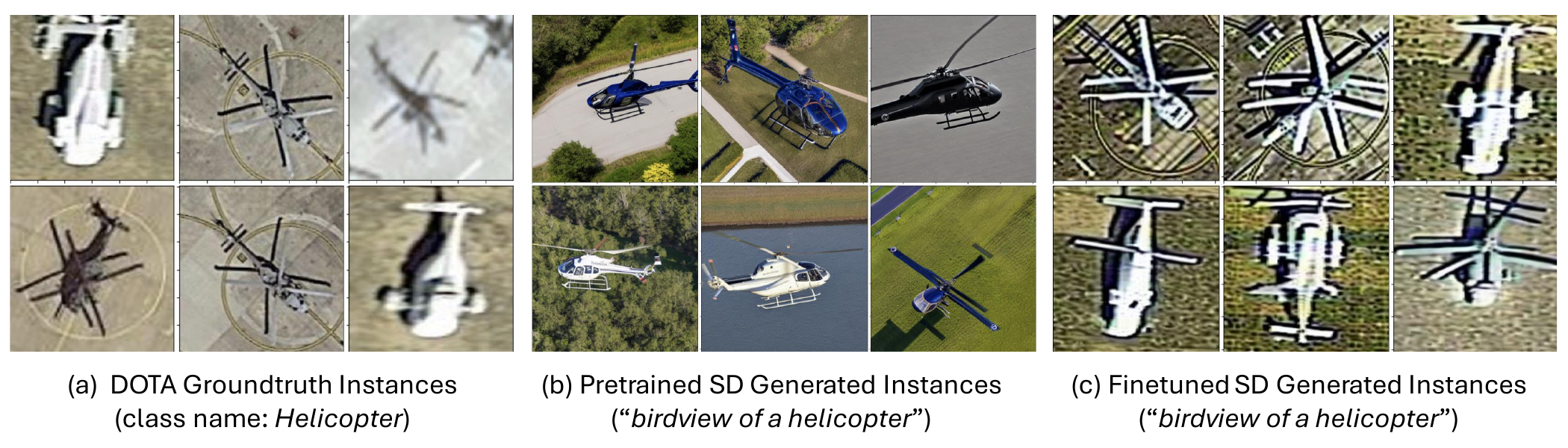}
    \caption{Comparison of (a) aerial imagery object instances; (b) off-the-shelf pretrained SD model generation results; (c) our LoRA finetuned SD model generation results.}
    \label{fig:finetune}
\end{figure}

StableDiffusion pretrained models are trained on mostly ground-level natural images, which lack exposure to the nuances of aerial-specific attributes (e.g., resolution loss, overhead perspective, etc.) As a result, using the off-the-shelf SD model for generating aerial objects/imagery usually performs poorly. As shown in Figure ~\ref{fig:finetune} (b), pre-trained SD's generated images of "helicopter" look drastically different from the original training set instances (Figure ~\ref{fig:finetune} (a)). Meanwhile, the helicopter instance's boundary does not align with the image boundary which will cause inaccurate instance coordinate extraction, potentially hurting the following detector training performance. Such difference emphasizes the need for domain-specific SD model fine-tuning and adaptation.

\vspace{-3mm}
\paragraph{Finetuning Strategy} We use the Low-Rank Adaptation (LORA) method~\cite{hu2021lora} to conduct efficient and effective SD model finetuning. The rationale behind this is that most concepts lying in aerial datasets are also present in the SD training set, such as helicopters, airplanes, vehicles, etc. Therefore, we only need to finetune the SD model to re-align the image semantics with aerial-specific viewpoints, appearances, etc., for which finetuning a small set of parameters should suffice. Meanwhile, LORA also requires a much smaller data volume which alleviates the need for large-scale dataset collection. 

In our experimentation, we sample a fixed amount of images per category to balance the training set and improve model training efficiency. To ensure the quality of generated images, we use uniform sampling but with a constraint of minimum object size. Any crops below 15x15 will not be sampled for training, and we take only at most 200 crops per category to assemble our finetuning dataset, which we found were able to yield satisfactory finetuning results.

\subsection{Object Synthesis and Copy-Paste Composition}

After finetuning, we then infer the SD model with the same prompt "birdview of <classname>" as the training process. With different random seeds, we could synthesize unlimited examples per category. Figure~\ref{fig:finetune} (c) shows the generated examples of "helicopter". As we can see, after SD finetuning, the generated instances align much better with the training datasets both in viewpoints and image styles. 

\vspace{-3mm}
\paragraph{Copy-Paste Synthetic Composition}
To use the synthetic object instances for the downstream object detection tasks, we use the Copy-Paste method~\cite{ghiasi2021simple} to conduct the final image composition and data augmentation. We randomly sample background images from the original training set, and our SD synthesized instances to paste onto the backgrounds. We calculate the box area and aspect ratio range per class from the original detection training dataset. When pasting the instances onto the background, we sample from the size/aspect ratio range based on the class's original distribution.
\section{Experiments}
\label{others}

\paragraph{Aerial Dataset}
We use DOTAv2.0 dataset \cite{9560031} throughout our experiments, which is one of the most representative aerial object detection datasets that contains 18 categories, such as airplane, small and large vehicles, swimming pool, etc.

We follow the standard procedure of processing a high-resolution dataset for detection task \cite{akyon2022sahi, Unel2019ThePO} by tiling each image into 512x512 patches with 200-pixel overlaps in between. We then train and evaluate our detectors on the tiled dataset.

\paragraph{Stable Diffusion Configuration}
For SD finetuning, we initialize the SD model with v1.5 pretrained weight \cite{Rombach_2022_CVPR}, and use the standard model config. We finetune it with batch size 1, the learning rate of 3e-4, and total iterations of 100k. VAE and CLIP modules are frozen during finetuning. 
For SD inference, we set the number of denoising steps to 50, guidance scale 7.5. We then use seed from 0 to 19 generating 200 different images per class.

\begin{figure}[!b]
\vspace{-3mm}
    \centering
    \includegraphics[width=\textwidth]{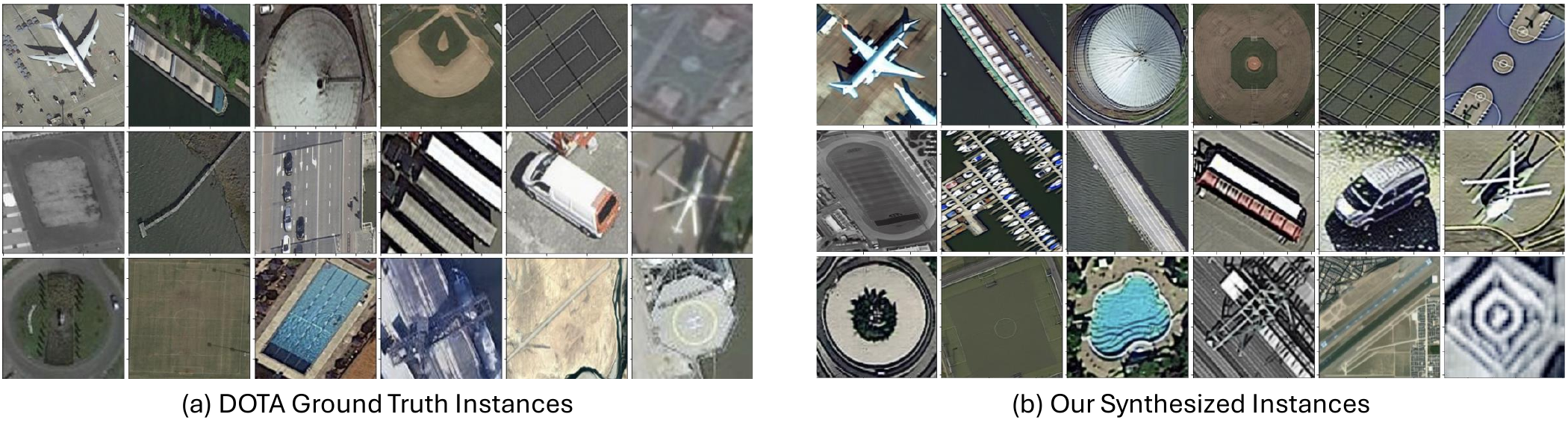}
    \caption{Comparison of (a) GT instances and (b) our SD synthesized instances.}
    \label{fig:allclass_results}
\end{figure}

\begin{figure}[!b]
    \centering
    \includegraphics[width=\textwidth]{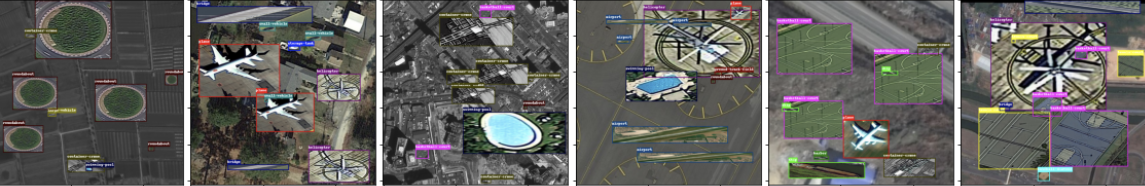}
    \caption{Composed synthetic images with the corresponding bboxes.}
    \vspace{-3mm}
    \label{fig:composed}
\end{figure}

\subsection{Synthetic Data Visualization}

Figure~\ref{fig:allclass_results} shows the comparison of DOTAv2.0 groundtruth instances and our finetuned SD model's synthesized instances for each class. For most classes (such as airplane, ship, storage-tank and swimming pool, etc.), the synthesized instances resemble the ground-truth instances in the DOTAv2.0 training samples well. For Copy-Paste augmentation, these SD-generated images are randomly sampled, rescaled, and pasted onto background images from the original dataset. Some examples are shown in Figure~\ref{fig:composed}. The composed synthetic images are used to augment the detector's training set.

We note that certain limitations exist in our Copy-Paste methods without having the semantic masks of the instances, thus the pasted image patch can have non-matching backgrounds. One promising solution is to leverage the off-the-shelf/finetuned segment-anything (SAM)~\cite{kirillov2023segment} model to further get the precise mask of the instances and then apply Copy-Paste, which we leave as future work.

\subsection{Object Detection Augmentation Results}
To evaluate our method's effectiveness, we conducted extensive data augmentation experiments on the object detection task using the DOTAv2.0 benchmark. Specifically, we tested both the two-stage detector, Faster-RCNN, and the single-stage detectors, YOLOV3 and RetinaNet. For backbone selection, we evaluated both CNN (ResNet-50) and transformer-based backbones (Swin-T and HorNet-T). For training data, our original (baseline) training set has 50k positive and 100k negative images, we start by mixing in 10k synthetic images (\textit{+10k}), then progress on mixing in 50k synthetic images (\textit{+50k}), to reach a balance of total positive and negative images in the detector training set. 

Due to the intrinsic nature of aerial images, the DOTAv2.0 dataset comes with many long-tail classes. In Table~\ref{tab:exp-results}, we sort the classes based on the number of images in the training set, in descending order from left to right. From "baseball-diamond" to "helipad" are what we consider long-tail classes, of which the total number of images per class is no greater than 200. All metrics in experiments are reported with mean average precision, \textit{mAP@(0.5:0.95)}.

\paragraph{Overall Experiment Results}

Table~\ref{tab:exp-results} shows the overall experimental results of our proposed approach across different backbones, detectors, and data augmentation settings. Both two-stage detectors like F-RCNN+SwinT backbone or F-RCNN+HorNet see immediate benefits of mixing in the synthetics, total AP gains range from \textbf{+1.2\%} to \textbf{+2.7\%}. Single-stage detectors also largely benefit from the synthetic data. Specifically, YOLOV3 saw a significant AP gain of \textbf{+3.6\%}, and RetinaNet+R50 backbone got a slight boost of \textbf{+1.6\%} mAP.

\paragraph{Higher Long-Tail Class Improvement} Meanwhile, we can also observe that most long-tail classes generally got more obvious AP boosts from our synthetic mix-ins. 
Based on Table~\ref{tab:exp-results}, the total gains of general classes (from small-vehicle to swimming pool) across models is on average \textbf{+1.4\%} in AP, whereas the total gains of long-tail classes (from baseball-diamond to helipad) have an average AP boost of \textbf{+4.1\%} points, across models. Especially for the "helipad" class, F-RCNN with HorNet backbone shows a \textbf{+30.3\%} mAP boost.

\begin{table}[!t]
\caption{Synthetic Augmentation Performance across Different Detector Architectures.}
\label{tab:exp-results}
\renewcommand\arraystretch{1.5}
\resizebox{\textwidth}{!}{%
\begin{tabular}{@{}lll|lllllllll|lllllllll@{}}
\toprule
\multicolumn{1}{c}{\textbf{Model}} & \multicolumn{1}{c}{\textbf{\begin{tabular}[c]{@{}c@{}}Train\\ Set\end{tabular}}} & \textbf{mAP} & \textbf{\begin{tabular}[c]{@{}l@{}}small\\ -vehicle\end{tabular}} & \textbf{\begin{tabular}[c]{@{}l@{}}large\\ -vehicle\end{tabular}} & \textbf{ship} & \textbf{plane} & \textbf{harbor} & \textbf{bridge} & \textbf{\begin{tabular}[c]{@{}l@{}}tennis\\ -court\end{tabular}} & \textbf{\begin{tabular}[c]{@{}l@{}}storage\\ -tank\end{tabular}} & \textbf{\begin{tabular}[c]{@{}l@{}}swim-\\ ming\\ -pool\end{tabular}} & \textbf{\begin{tabular}[c]{@{}l@{}}base-\\ ball-\\ diam\end{tabular}} & \textbf{\begin{tabular}[c]{@{}l@{}}round\\ -about\end{tabular}} & \textbf{\begin{tabular}[c]{@{}l@{}}ground-\\ track-\\ field\end{tabular}} & \textbf{\begin{tabular}[c]{@{}l@{}}soccer-\\ ball-\\ field\end{tabular}} & \textbf{\begin{tabular}[c]{@{}l@{}}basket\\ -ball\\ -court\end{tabular}} & \textbf{\begin{tabular}[c]{@{}l@{}}air-\\ port\end{tabular}} & \textbf{\begin{tabular}[c]{@{}l@{}}heli-\\ copter\end{tabular}} & \textbf{\begin{tabular}[c]{@{}l@{}}conta-\\ iner\\ -crane\end{tabular}} & \textbf{\begin{tabular}[c]{@{}l@{}}heli-\\ pad\end{tabular}} \\ \midrule
\multirow{3}{*}{\begin{tabular}[c]{@{}l@{}}F-RCNN\\ + SwinT\end{tabular}} & baseline & 0.358 & 0.238 & 0.53 & 0.53 & 0.596 & 0.388 & 0.219 & 0.769 & 0.374 & 0.253 & 0.327 & 0.352 & \textbf{0.49} & \textbf{0.366} & 0.381 & 0.203 & 0.288 & 0.002 & 0.135 \\ \cmidrule(l){2-21} 
 & +10k & 0.362 & \textbf{0.24} & \textbf{0.534} & 0.53 & 0.595 & \textbf{0.394} & \textbf{0.222} & 0.762 & \textbf{0.383} & 0.255 & 0.327 & \textbf{0.363} & 0.488 & 0.325 & 0.368 & 0.214 & 0.286 & 0.004 & 0.236 \\ \cmidrule(l){2-21} 
 & +50k & \textbf{0.37} & 0.239 & 0.526 & \textbf{0.532} & 0.596 & 0.386 & 0.211 & \textbf{0.773} & 0.375 & \textbf{0.268} & \textbf{0.345} & 0.362 & 0.487 & 0.33 & \textbf{0.416} & \textbf{0.227} & \textbf{0.306} & \textbf{0.005} & \textbf{0.269} \\ \midrule
 &  &  &  &  &  &  &  &  &  &  &  &  &  &  &  &  &  &  &  &  \\ \midrule
\multirow{3}{*}{\begin{tabular}[c]{@{}l@{}}F-RCNN\\ +HorNet\end{tabular}} & baseline & 0.337 & 0.229 & \textbf{0.519} & 0.513 & 0.585 & 0.393 & 0.201 & 0.77 & 0.371 & \textbf{0.266} & 0.336 & 0.329 & 0.444 & 0.305 & 0.319 & 0.189 & \textbf{0.301} & 0.003 & 0 \\ \cmidrule(l){2-21} 
 & +10k & 0.351 & 0.233 & 0.507 & 0.525 & \textbf{0.596} & 0.395 & 0.198 & 0.761 & \textbf{0.379} & 0.265 & 0.327 & 0.346 & \textbf{0.463} & \textbf{0.34} & \textbf{0.43} & \textbf{0.237} & 0.283 & \textbf{0.021} & 0.013 \\ \cmidrule(l){2-21} 
 & +50k & \textbf{0.364} & \textbf{0.239} & 0.515 & \textbf{0.527} & 0.587 & \textbf{0.401} & \textbf{0.208} & \textbf{0.776} & 0.369 & 0.253 & \textbf{0.336} & \textbf{0.346} & 0.46 & 0.317 & 0.398 & 0.229 & 0.277 & 0.005 & \textbf{0.303} \\ \midrule
 &  &  &  &  &  &  &  &  &  &  &  &  &  &  &  &  &  &  &  &  \\ \midrule
\multirow{3}{*}{YOLOV3} & baseline & 0.209 & \textbf{0.198} & 0.304 & 0.428 & 0.447 & 0.18 & 0.122 & 0.581 & 0.321 & \textbf{0.171} & 0.166 & 0.208 & 0.223 & 0.072 & 0.149 & 0.002 & 0.127 & \textbf{0.007} & \textbf{0.05} \\ \cmidrule(l){2-21} 
 & +10k & 0.211 & 0.195 & 0.372 & 0.45 & 0.485 & 0.138 & \textbf{0.135} & 0.377 & 0.359 & 0.123 & 0.214 & 0.224 & 0.179 & 0.088 & \textbf{0.253} & \textbf{0.022} & \textbf{0.14} & 0.001 & 0.034 \\ \cmidrule(l){2-21} 
 & +50k & \textbf{0.245} & 0.196 & \textbf{0.401} & \textbf{0.485} & \textbf{0.516} & \textbf{0.211} & 0.121 & \textbf{0.6} & \textbf{0.364} & 0.157 & \textbf{0.253} & \textbf{0.294} & \textbf{0.307} & \textbf{0.12} & 0.247 & 0.015 & 0.118 & 0.002 & 0.003 \\ \midrule
 &  &  &  &  &  &  &  &  &  &  &  &  &  &  &  &  &  &  &  &  \\ \midrule
\multirow{3}{*}{\begin{tabular}[c]{@{}l@{}}RetinaNet\\ +R50\end{tabular}} & baseline & 0.237 & 0.165 & 0.344 & 0.429 & \textbf{0.515} & 0.245 & 0.16 & 0.672 & 0.307 & 0.19 & 0.24 & 0.279 & 0.314 & 0.085 & 0.251 & 0.009 & 0.053 & 0 & 0 \\ \cmidrule(l){2-21} 
 & +10k & \textbf{0.253} & 0.167 & \textbf{0.359} & 0.433 & 0.502 & \textbf{0.256} & 0.164 & 0.689 & \textbf{0.317} & \textbf{0.196} & \textbf{0.274} & \textbf{0.299} & 0.332 & \textbf{0.095} & 0.29 & \textbf{0.068} & 0.113 & 0 & 0 \\ \cmidrule(l){2-21} 
 & +50k & 0.25 & \textbf{0.169} & 0.358 & \textbf{0.435} & 0.498 & 0.253 & \textbf{0.165} & \textbf{0.7} & 0.304 & 0.191 & 0.261 & 0.288 & \textbf{0.333} & 0.094 & \textbf{0.309} & 0.013 & \textbf{0.134} & 0 & \textbf{0.001} \\ \bottomrule
\end{tabular}%
}
\end{table}

\paragraph{Compare with Vanilla CopyPaste}
We then compare our method with vanilla CopyPaste augmentation on the long-tail classes \cite{ghiasi2021simple} in Table~\ref{tab:copy-paste-aug}. Since we don't have ground-truth segmentation masks, the Copy-Paste Augmentation operates on bbox-level. We conduct two experiments: (1) the default configuration of Copy-Pasting regardless of classes (\textit{aug-all-class}); (2) Apply our synthetic Copy-Paste method with the helipad instances (\textit{aug-helipad}). Helipad has the least number of images in DOTA-v2.0 training set (only with 91 images), compared to small-vehicle having 24,341 images present in the training set, thus it's especially hard to improve AP in this class. 

Based on Table~\ref{tab:copy-paste-aug}'s results, we can see that vanilla Copy-Paste augmentation drags down the overall mAP since there are no new instances involved, while our Copy-Paste method on synthetic helipad instances shows obvious gain on this specific hard-case class, further indicating our synthesized objects brings new information to the training dataset. 

\paragraph{Ablation Study of Sampling Strategies}
We experimented with two strategies when sampling dense regions for SD+LoRA finetune. The results are shown in Table~\ref{tab:ablate-SD-training-data-sampling}. (1) \textit{uniform-sample}: We start by uniform sampling the cropped dense regions regardless of size, as a result, we can see extremely low-resolution images being sampled into the finetuning set, eg. 32x32-sized helicopters and 10x10-sized helipads. (2) \textit{min-resolution control}: We then add a constraint to the minimum resolution when sampling. regions smaller than 15x15 are ignored during sampling. 
We finetune SD separately on these two settings. Based on Table~\ref{tab:ablate-SD-training-data-sampling}, we can see that the training detector on SD finetuned with controlling minimum resolution shows better performance on long-tail classes.

\begin{table}[!tb]
\caption{Baseline CopyPaste vs Long-Tail Augmentation Results}
\label{tab:copy-paste-aug}
\resizebox{\textwidth}{!}{%
\begin{tabular}{lll|lllllllll|lllllllll}
\toprule
\multicolumn{1}{c}{\textbf{Model}} & \multicolumn{1}{c}{\textbf{\begin{tabular}[c]{@{}c@{}}Train\\ Set\end{tabular}}} & \textbf{mAP} & \textbf{\begin{tabular}[c]{@{}l@{}}small\\ -vehicle\end{tabular}} & \textbf{\begin{tabular}[c]{@{}l@{}}large\\ -vehicle\end{tabular}} & \textbf{ship} & \textbf{plane} & \textbf{harbor} & \textbf{bridge} & \textbf{\begin{tabular}[c]{@{}l@{}}tennis\\ -court\end{tabular}} & \textbf{\begin{tabular}[c]{@{}l@{}}storage\\ -tank\end{tabular}} & \textbf{\begin{tabular}[c]{@{}l@{}}swim-\\ ming\\ -pool\end{tabular}} & \textbf{\begin{tabular}[c]{@{}l@{}}base-\\ ball-\\ diam\end{tabular}} & \textbf{\begin{tabular}[c]{@{}l@{}}round\\ -about\end{tabular}} & \textbf{\begin{tabular}[c]{@{}l@{}}ground-\\ track-\\ field\end{tabular}} & \textbf{\begin{tabular}[c]{@{}l@{}}soccer-\\ ball-\\ field\end{tabular}} & \textbf{\begin{tabular}[c]{@{}l@{}}basket\\ -ball\\ -court\end{tabular}} & \textbf{\begin{tabular}[c]{@{}l@{}}air-\\ port\end{tabular}} & \textbf{\begin{tabular}[c]{@{}l@{}}heli-\\ copter\end{tabular}} & \textbf{\begin{tabular}[c]{@{}l@{}}conta-\\ iner\\ -crane\end{tabular}} & \textbf{\begin{tabular}[c]{@{}l@{}}heli-\\ pad\end{tabular}} \\ \hline
\multirow{3}{*}{\begin{tabular}[c]{@{}l@{}}F-RCNN\\ +SwinT\end{tabular}}  & \begin{tabular}[c]{@{}l@{}}base- \\ line\end{tabular} & \textbf{0.358} & 0.238 & \textbf{0.53} & 0.53 & 0.596 & 0.388 & \textbf{0.219} & \textbf{0.769} & \textbf{0.374} & 0.253 & 0.327 & \textbf{0.352} & \textbf{0.49} & \textbf{0.366} & \textbf{0.381} & 0.203 & \textbf{0.288} & 0.002 & 0.135 \\ \cline{2-21} 
& \begin{tabular}[c]{@{}l@{}}aug, \\ all cls\end{tabular} & 0.353 & \textbf{0.24} & 0.521 & \textbf{0.531} & \textbf{0.598} & \textbf{0.39} & 0.212 & \textbf{0.769} & 0.371 & \textbf{0.262} & \textbf{0.351} & 0.342 & 0.483 & 0.33 & 0.357 & \textbf{0.209} & 0.287 & \textbf{0.009} & 0.101 \\ \cline{2-21} 
 & \begin{tabular}[c]{@{}l@{}}aug, \\ helipad\end{tabular} & 0.354 & 0.239 & 0.517 & 0.524 & 0.596 & 0.378 & 0.207 & 0.754 & \textbf{0.374} & 0.254 & 0.337 & 0.343 & 0.475 & 0.318 & 0.34 & 0.167 & 0.272 & 0.007 & \textbf{0.269} \\ \hline
\end{tabular}%
}
\vspace{-3mm}
\end{table}

\begin{table}[!tb]
\caption{Ablation of Sampling Strategy}
\label{tab:ablate-SD-training-data-sampling}
\resizebox{\textwidth}{!}{%
\begin{tabular}{@{}llc|ccccccccc|ccccccccc@{}}
\toprule
\multicolumn{1}{c}{\textbf{Model}} & \multicolumn{1}{c}{\textbf{\begin{tabular}[c]{@{}c@{}}Train\\ Set\end{tabular}}} & \multicolumn{1}{l|}{\textbf{mAP}} & \multicolumn{1}{l}{\textbf{\begin{tabular}[c]{@{}l@{}}small\\ -vehicle\end{tabular}}} & \multicolumn{1}{l}{\textbf{\begin{tabular}[c]{@{}l@{}}large\\ -vehicle\end{tabular}}} & \multicolumn{1}{l}{\textbf{ship}} & \multicolumn{1}{l}{\textbf{plane}} & \multicolumn{1}{l}{\textbf{harbor}} & \multicolumn{1}{l}{\textbf{bridge}} & \multicolumn{1}{l}{\textbf{\begin{tabular}[c]{@{}l@{}}tennis\\ -court\end{tabular}}} & \multicolumn{1}{l}{\textbf{\begin{tabular}[c]{@{}l@{}}storage\\ -tank\end{tabular}}} & \multicolumn{1}{l|}{\textbf{\begin{tabular}[c]{@{}l@{}}swim-\\ ming\\ -pool\end{tabular}}} & \multicolumn{1}{l}{\textbf{\begin{tabular}[c]{@{}l@{}}base-\\ ball-\\ diam\end{tabular}}} & \multicolumn{1}{l}{\textbf{\begin{tabular}[c]{@{}l@{}}round\\ -about\end{tabular}}} & \multicolumn{1}{l}{\textbf{\begin{tabular}[c]{@{}l@{}}ground-\\ track-\\ field\end{tabular}}} & \multicolumn{1}{l}{\textbf{\begin{tabular}[c]{@{}l@{}}soccer-\\ ball-\\ field\end{tabular}}} & \multicolumn{1}{l}{\textbf{\begin{tabular}[c]{@{}l@{}}basket\\ -ball\\ -court\end{tabular}}} & \multicolumn{1}{l}{\textbf{\begin{tabular}[c]{@{}l@{}}air-\\ port\end{tabular}}} & \multicolumn{1}{l}{\textbf{\begin{tabular}[c]{@{}l@{}}heli-\\ copter\end{tabular}}} & \multicolumn{1}{l}{\textbf{\begin{tabular}[c]{@{}l@{}}conta-\\ iner\\ -crane\end{tabular}}} & \multicolumn{1}{l}{\textbf{\begin{tabular}[c]{@{}l@{}}heli-\\ pad\end{tabular}}} \\ \midrule
\multirow{3}{*}{\begin{tabular}[c]{@{}l@{}}F-RCNN\\ + swinT\end{tabular}} & \begin{tabular}[c]{@{}l@{}} base-\\ line\end{tabular} & 0.358 & 0.238 & 0.53 & 0.53 & 0.596 & 0.388 & \textbf{0.219} & 0.769 & 0.374 & 0.253 & 0.327 & 0.352 & \textbf{0.49} & \textbf{0.366} & 0.381 & 0.203 & 0.288 & 0.002 & 0.135 \\ \cmidrule(l){2-21} 
 & \begin{tabular}[c]{@{}l@{}} uniform\\ sample\end{tabular} & 0.366 & \textbf{0.24} & \textbf{0.533} & 0.531 & 0.596 & \textbf{0.395} & 0.217 & \textbf{0.773} & \textbf{0.375} & \textbf{0.27} & 0.33 & 0.346 & 0.485 & 0.329 & 0.378 & 0.198 & 0.251 & 0.002 & \textbf{0.337} \\ \cmidrule(l){2-21} 
  & \begin{tabular}[c]{@{}l@{}} min-res\\ control\end{tabular} & \textbf{0.37} & 0.239 & 0.526 & \textbf{0.532} & 0.596 & 0.386 & 0.211 & \textbf{0.773} & \textbf{0.375} & 0.268 & \textbf{0.345} & \textbf{0.362} & 0.487 & 0.33 & \textbf{0.416} & \textbf{0.227} & \textbf{0.306} & \textbf{0.005} & 0.269 \\  \bottomrule
\end{tabular}}
\vspace{-3mm}
\end{table}

\section{Discussion \& Future Work}

The current SD finetuning approach generates synthetic training images that brings new knowledge to detection model, as a result, mixing those images into original training dataset boosts detector mAP scores especially on long-tail classes. However we've also tried using only the synthetic images for training, without mixing in the original images from DOTAv2.0. We noticed the performance is extremely low, almost all classes got a close-to-zero mAP score. This may due to the fact that the synthetic images were composed without masks, detector may overfit to the boundries of the bboxes when trained only on those images. For the next step, we can try zero-shot segmentation methods to generate object masks, then compose synthetic images to eliminate current bbox boundries.

\section{Conclusion}
Aerial object detection task faces extensive dataset limitations like data collection constraints and skewed class distributions. Synthetic data, especially the recent diffusion-based techniques, emerged as a promising way to alleviate these constraints. Nevertheless, a direct application to aerial images is proved suboptimal, highlighting the need for a domain-specific approach. Our proposed framework, integrating sparse-to-dense semantic ROI extraction, LORA fine-tuning, and the Copy-Paste technique, presents a synergistic solution tailored to the unique challenges of aerial object detection. Extensive experiments demonstrate the effectiveness of our proposed synthetic framework, paving the way for more accurate and comprehensive aerial scene analyses.

\bibliographystyle{plain}
\bibliography{diffusion}

\end{document}